\documentclass{article}
\usepackage{arxiv}

\usepackage[utf8]{inputenc} 
\usepackage[T1]{fontenc}    
\usepackage{hyperref}       
\usepackage{url}            
\usepackage{booktabs}       
\usepackage{framed,multirow}
\usepackage[bottom]{footmisc}
\usepackage{amssymb}
\usepackage{latexsym}
\usepackage{subcaption}
\usepackage{url}
\usepackage{graphicx}
\usepackage{multirow}
\usepackage{booktabs}
\usepackage{tabularx}
\usepackage{placeins}
\usepackage[style=numeric, sorting=none]{biblatex} 
\usepackage{url}
\usepackage{xcolor}
\definecolor{newcolor}{rgb}{.8,.349,.1}
\usepackage{amsmath}
\usepackage{hyperref}
\title{Adaptive Adversarial Augmentation for Controllable Face Synthesis}

\author{
Saransh Suri$^{1}$\thanks{Work done while at Indian Institute of Technology Jodhpur.}
\quad
Shivang Agarwal$^{2}$\thanks{Corresponding author.}
\quad
Mayank Vatsa$^{3}$
\quad
Richa Singh$^{3}$
\\[4pt]
$^{1}$Manipal University Jaipur, India\\
$^{2}$Birla Institute of Technology and Science Pilani Dubai Campus, Dubai, United Arab Emirates\\
$^{3}$Indian Institute of Technology Jodhpur, India
}

\renewcommand{\shorttitle}{Controllable Face Synthesis}

\hypersetup{
pdftitle={A template for the arxiv style},
pdfsubject={q-bio.NC, q-bio.QM},
pdfauthor={David S.~Hippocampus, Elias D.~Striatum},
pdfkeywords={First keyword, Second keyword, More},
}

\begin{document}
\maketitle

\begin{abstract}
Synthetic data provides a scalable alternative to real-world datasets for training face recognition models, particularly under challenging conditions such as low resolution, occlusion, and masks. Yet, most approaches lack diversity and fail to generalize effectively. We propose Ensemble Feedback Controllable Synthesis (EFCS), a guided framework that generates diverse and challenging samples while preserving visual realism. EFCS expands distributional variability, often reflected in higher FID and KID scores compared to single-feedback and random synthesis, while maintaining high precision. Recognition models trained on EFCS data consistently outperform baselines across multiple benchmarks, showing improved generalization to real-world scenarios. Furthermore, we introduce an analytically motivated formulation linking perturbation-induced difficulty, sample utility, and performance degradation, offering principled insights into balancing synthetic data complexity for optimal training. Together, these contributions establish EFCS as an effective and analytically grounded approach for bridging the gap between synthetic and real datasets.
\end{abstract}


\section{Introduction}
\label{sec:intro}

Face recognition (FR) systems have witnessed significant advancements, largely driven by the widespread availability of large-scale identity-labeled face datasets. Nevertheless, this rapid progress has concomitantly raised substantial concerns related to privacy, ethics, and inherent demographic biases arising from real identity data \cite{bias-diverisity-synthetic_2024_WACV,bias-cvpr,legal-tifs21_privacy,regulation2016regulation_gdpr_privacy}. To address these critical issues, synthetic face datasets have emerged as an attractive alternative for training FR models, offering considerable potential for privacy protection and bias mitigation \cite{synthetic-survey,dabouei2018fastgeometricallyperturbedadversarialfaces,SynFace}. Despite these promising attributes, FR models trained exclusively on synthetic data consistently underperform relative to those trained on real-world datasets \cite{human_synthetic}. This performance gap primarily arises from two fundamental challenges: insufficient intra-class variations within synthetic images and pronounced domain mismatches between synthetic and real data distributions \cite{wu2025vec2facescalingfacedataset,SynFace,bias-diverisity-synthetic_2024_WACV}. These limitations adversely affect the generalization capabilities of FR models, particularly in complex scenarios such as low-resolution surveillance images \cite{sface}, faces with occlusions or injuries \cite{chiranjeev2025beyond}, and cross-domain selfie versus ID verification \cite{selfie}, as highlighted in recent analyses of synthetic data utility and generalization~\cite{synthetic-survey,bias-diverisity-synthetic_2024_WACV}.These observations motivate the need for synthetic data generation strategies
that explicitly optimize training utility rather than visual fidelity alone.

\begin{figure}[t] 
    \centering
    \includegraphics[width=0.8\columnwidth]{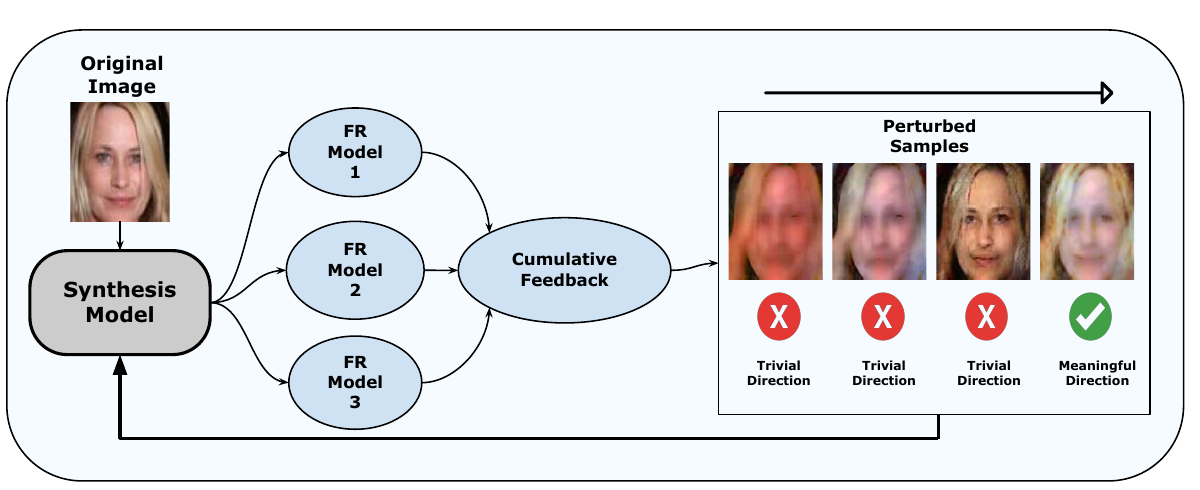}
    \caption{Illustration highlighting the necessity of adaptive synthesis. Controlled perturbations (hard samples) introduce targeted complexity to synthetic data, enhancing model robustness beyond traditional synthesis methods.}
\end{figure}

Current approaches for synthetic face generation exhibit significant limitations. Generative Adversarial Networks (GANs), despite their popularity, frequently fail to produce adequate intra-class diversity, resulting in synthetic datasets characterized by limited variations within individual identity classes \cite{ganforsyn_synthetic, Melzi_2023_ICCV_GandiffFace, faceid-gan_identity}. This paucity of diversity often leads FR models trained on GAN-generated datasets to suffer from overfitting and reduced generalization. On the other hand, recent diffusion-based synthesis methods generate high-quality images but lack the fine-grained control necessary to produce targeted variations or precise manipulations of facial attributes \cite{FG25, kawar, DCFace, Liu_2023_WACV_Guidance_classifier}. Moreover, existing methods generally overlook the critical domain gap, hindering effective transferability of synthetically-trained models to practical real-world applications \cite{bias-cvpr, bias-diverisity-synthetic_2024_WACV, human_synthetic, surbhi}.

\begin{figure*}[!t]
    \centering
    \includegraphics[width=\textwidth]{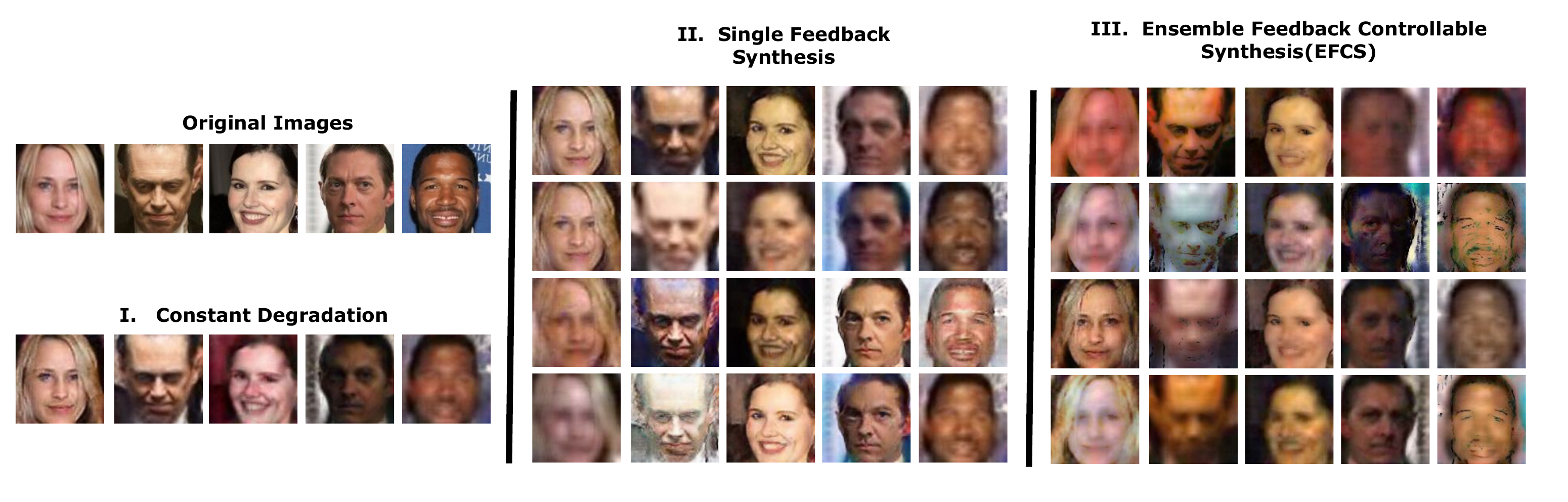}
    \caption{Qualitative comparison of synthetic samples: (I) constant degradation, (II) single feedback synthesis~\cite{CFSM}, and (III) proposed EFCS method.}
    \label{fig:grid}
\end{figure*}

To address these persistent challenges, this paper introduces a novel approach termed \emph{Ensemble Feedback Controllable Synthesis (EFCS)}. EFCS guides the synthetic data generation process by leveraging ensemble feedback from multiple pre-trained FR models, thus producing synthetic face images tailored specifically to enhance model training. The generated images are rich in intra-class and inter-class diversity. By generating diverse yet realistically challenging synthetic samples through controlled feedback mechanisms, EFCS enhances the robustness and generalization of FR models. Crucially, as shown in Figure \ref{fig:grid}, EFCS ensures that the generated samples remain appropriately challenging without introducing excessive difficulty that could cause the models to overfit irrelevant attributes, such as background or clothing \cite{adaface}. 

Our contributions are threefold:
\begin{itemize}
    \item We empirically demonstrate that FR models trained solely on synthetic data without guided feedback consistently fall short of real-data model performance, highlighting the necessity of a more sophisticated synthetic data generation strategy.
    \item We propose the \emph{Ensemble Feedback Controllable Synthesis (EFCS)}, a novel guided synthesis framework leveraging feedback from an ensemble of FR models to produce highly beneficial synthetic training samples.
    \item We introduce a theoretical formulation that quantifies the relationship between perturbation-induced difficulty, sample utility, and performance degradation. This formulation provides insights into balancing the complexity of synthetic data for optimal FR model training.
\end{itemize}

\section{Literature Review}
\label{sec:Literature Review}

Controllable face synthesis methods \cite{CFSM, kawar, Yu_2023_ICCV_TextGuided} have emerged as particularly promising avenues for generating diverse and high-quality synthetic face datasets. These methods offer fine-grained control over facial attributes via manipulation in the latent style space. Such attribute control is motivated primarily by the need to address privacy concerns linked to the use of real, identity-labeled datasets and to manage challenges arising from noisy labels and imbalanced demographic representation.

Among various synthesis methods, GANs have been central to synthetic face image generation \cite{IJCB23-synthdistill, ganforsyn_synthetic, unsupervised_synthetic}. For instance, SFace \cite{sface} employs StyleGAN2-ADA to create synthetic datasets closely aligned with real images through a 1:1 correspondence strategy. Despite this alignment, experimental evaluations demonstrate performance gaps, often exceeding 25\% in verification accuracy, between models trained on synthetic data and those trained on real images. This gap highlights a significant limitation: synthetic data frequently fails to replicate the complexity, diversity, and subtle intra-class variations inherent to real-world images. To mitigate this, several approaches have been explored. SynFace \cite{SynFace} introduces a method of mixing intermediate latent states from synthetic identities to enhance intra-class diversity. Nevertheless, SynFace predominantly generates frontal-view faces, necessitating additional real-image mixing to adequately bridge the persistent domain gap between synthetic and real-world face images.

Recently, diffusion-based models have garnered increased attention as viable alternatives to GAN-based methods \cite{DCFace, Liu_2023_WACV_Guidance_classifier, IdentityPreservingSynthesis}. Diffusion models enable precise generation and manipulation of facial attributes guided by diverse style-based prompts, including text descriptions and sketches \cite{Hasan_2024_WACV_textGuided}. Despite providing greater control over specific facial features, diffusion models still face substantial challenges ~\cite{wu2025vec2facescalingfacedataset}. Primarily, they struggle to produce synthetic samples sufficiently diverse and challenging to robustly train FR models, thereby limiting the practical utility of generated datasets \cite{demodalize_synthetic}.

Despite significant methodological advancements, existing synthetic face generation techniques commonly encounter critical shortcomings in complex FR tasks, such as low-resolution surveillance face recognition, face recognition under varying pose conditions, and identification scenarios involving partially occluded or injured faces. The key limitations consistently observed across current methodologies include: (1) inadequate intra-class diversity within synthetic datasets, (2) enduring domain mismatches between synthetic and real-world image distributions, and (3) limited capability to generate sufficiently challenging samples that robustly enhance FR model training \cite{cvpr20-synthetic}.


\section{Ensemble Feedback Controllable Synthesis}

\begin{figure}[t]
\centering
\includegraphics[width=\linewidth]{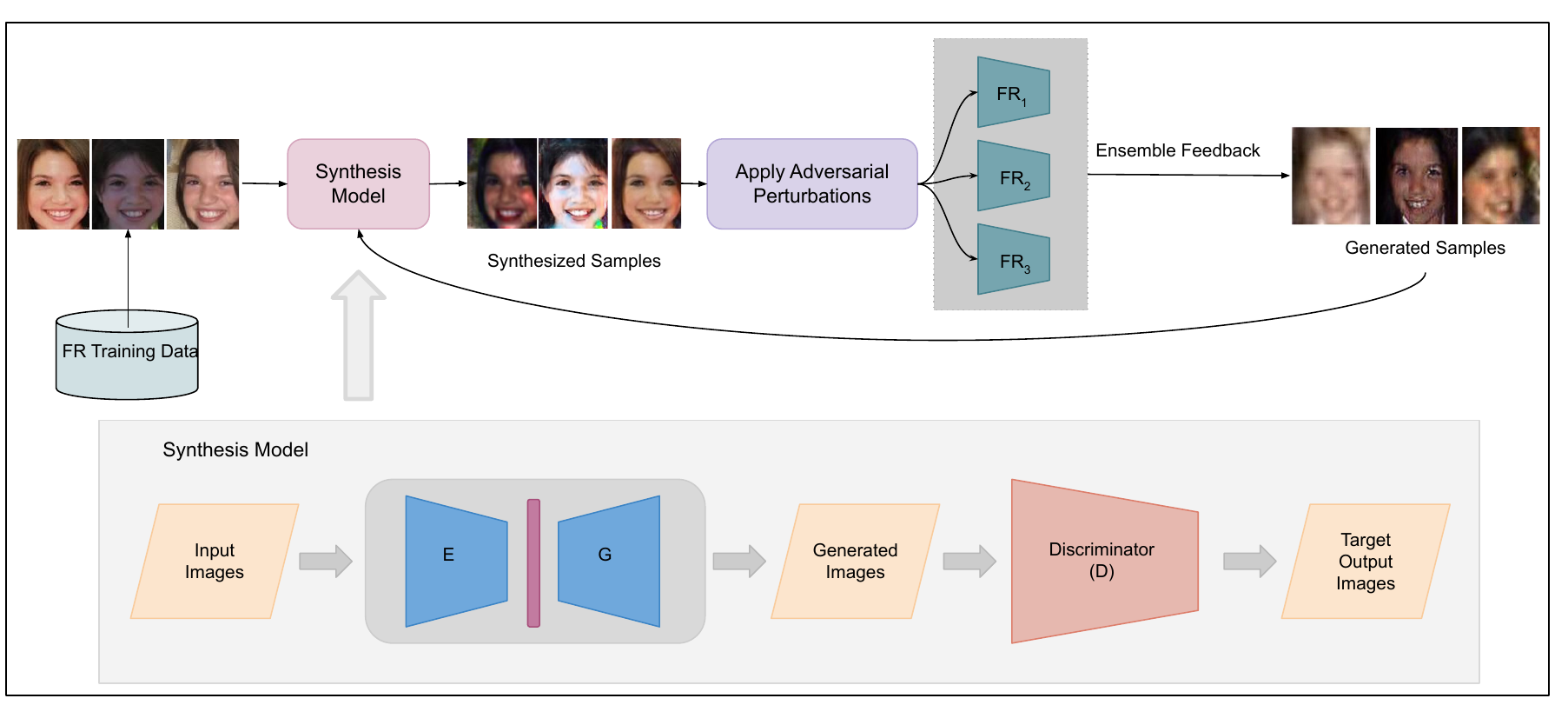}
\caption{Overview of the proposed EFCS framework showing synthesis, perturbation, and ensemble feedback.}
\label{arch}
\end{figure}
Given an original face image $x$ with true identity label $y$, the primary goal is to generate an adversarially perturbed synthetic face image $x_{perturb}$ that significantly increases the difficulty of correct recognition by FR models, while maintaining realism and identity preservation. Unlike adversarial attacks designed to expose worst-case vulnerabilities at inference time, EFCS uses adversarial optimization as a training-time augmentation mechanism. The objective is not robustness against malicious perturbations, but controlled difficulty injection that improves representation learning. This goal is formally expressed as:

\begin{equation}
    x_{perturb} = G(x, {perturb})
\end{equation}

where $G(\cdot)$ denotes the controllable face synthesis model, and $perturb$ represents the perturbation drawn from a structured latent space. Unlike random perturbations, this structured perturbation ensures semantic coherence and realistic transformations.

Let us consider an ensemble of $N$ pre-trained FR models, denoted as $\{f_1, f_2, \dots, f_N\}$. Each model $f_i$ has an associated loss function $L_i(f_i(x), y)$ that quantifies its performance in recognizing image $x$ as identity $y$. Typical loss functions include cross-entropy or margin-based verification losses (e.g., ArcFace \cite{ArcFace}, AdaFace \cite{adaface}).

The aggregated recognition difficulty posed by the ensemble on an image $x$ is defined as the ensemble recognition loss:

\begin{equation}
x_{\text{ens}} =
\arg\max_{i \in \{1,\dots,N\}}
L_i\big(f_i(G(x,\text{perturb}_i)), y\big)
\end{equation}

Our adversarial objective is to maximize this ensemble recognition loss, thus explicitly making the synthesized image harder to recognize:

\begin{equation}
    \max_{perturb} L_{\text{ens}}(G(x, perturb), y)
\end{equation}

To ensure controlled and meaningful perturbations, we introduce a perturbation depth parameter $d$ that dynamically scales with the difficulty level of the input image as determined by the ensemble:

\begin{equation}
    d = \gamma \cdot L_{\text{ens}}(x, y)
\end{equation}

where $\gamma$ is a hyperparameter scaling factor. This depth controls the intensity of the perturbation, ensuring that harder samples (as per ensemble loss) receive appropriately stronger perturbations.

The perturbation vector $perturb$ is sampled from a structured distribution conditioned on the perturbation depth $d$:

\begin{equation}
    perturb^{(d)} \sim \mathcal{P}(d)
\end{equation}

To formally measure and maximize the training utility of the synthesized adversarial images, we define a training utility function $U$, which aims for a targeted increase $\delta$ in ensemble recognition difficulty:

\begin{equation}
U(x_{perturb^{(d)}}) =
- \left|
L_{\text{ens}}(x_{perturb^{(d)}}, y)
-
\left(L_{\text{ens}}(x, y) + \delta\right)
\right|
\end{equation}

Maximizing $U$ ensures the generated synthetic images are neither too trivial nor excessively difficult, thereby optimally contributing to model training.

\subsection{Proposed EFCS Architecture}

The proposed EFCS method employs a sophisticated two-stage architecture to systematically generate and utilize adversarial synthetic images. The overall framework is illustrated in Figure \ref{arch}.

The controllable synthesis architecture used in EFCS follows the generator
design introduced in CFSM~\cite{CFSM}, which  builds upon the encoder–decoder
framework. In our framework, this architecture is adopted as the base generator to ensure stable identity-preserving synthesis.

The primary novelty of EFCS lies not in redesigning the generator architecture,
but in the feedback-driven training mechanism applied to it. Specifically,
EFCS introduces three key extensions beyond CFSM:
(i) retraining the synthesis model itself using feedback signals,
(ii) replacing single-model feedback with ensemble-based feedback from multiple
face recognition models, and
(iii) introducing a utility-driven objective that explicitly controls the
difficulty of generated samples.

\subsubsection{Stage 1: Controllable Synthesis Model}

In Stage 1, a controllable synthesis model $G$ is designed and trained to produce face images with explicitly controlled and interpretable variations. This model comprises three critical components:

\noindent\textbf{Encoder:} The encoder module captures identity-preserving content from input face image $x$. It employs reflective padding at the borders, convolutional layers for robust feature extraction, and stride-2 downsampling layers to progressively reduce spatial resolution. To further refine feature extraction, instance-normalized residual blocks are incorporated. These operations result in a compact and informative representation of essential identity features, discarding irrelevant attributes such as background details or varying illuminations.

\noindent\textbf{Decoder:} The decoder reconstructs realistic face images from the encoded content representations. It introduces a latent style vector, generated by a dedicated Multi-Layer Perceptron (MLP), to modulate intermediate activations via Adaptive Instance Normalization (AdaIN) layers. AdaIN enables granular and interpretable control over the reconstruction process, systematically introducing controlled perturbations. The decoder’s architecture includes residual blocks with AdaIN for style infusion, nearest-neighbor interpolation for efficient upsampling, and convolutional layers for fine-grained feature refinement. Finally, a Tanh activation layer ensures synthesized images remain within plausible and realistic pixel value ranges.

\noindent\textbf{Subspace Layer:} A specialized subspace layer ensures latent perturbations are diverse yet coherent. This layer enforces orthogonality constraints on latent dimensions, which correspond to interpretable and independent facial variations such as pose, illumination, expression, and occlusion. The orthogonality constraint prevents collapse to trivial perturbations, maintaining both diversity and semantic realism of the synthesized face images. Regularization techniques are applied during training to uphold this constraint, making the generated synthetic samples highly valuable and broadly applicable for FR model enhancement.

\subsubsection{Stage 2: Adversarial Augmentation with Ensemble Feedback}

In Stage 2, the controllable synthesis model $G$ is integrated into an adversarial data augmentation pipeline driven by ensemble-based feedback. The goal of this stage is to generate synthetic adversarial images dynamically during the FR model training process, systematically targeting and challenging the weaknesses of the ensemble models.

\noindent\textbf{Latent Initialization:} Initially, a latent perturbation vector $perturb$ is randomly sampled from the structured perturbation space. This vector dictates the initial set of transformations applied to the input face image, defining the baseline for subsequent optimization.

\noindent\textbf{Feedback Loop and PGD Optimization:} Ensemble feedback is systematically used through Projected Gradient Descent (PGD). For each iteration, the current synthetic image is evaluated by all ensemble models to compute their respective recognition losses. These computed losses are then used to iteratively update the latent perturbation vector $perturb$. Specifically, PGD iteratively updates the latent perturbation so as to increase the recognition loss of each ensemble model, after which the perturbation is refined through the ensemble feedback optimization loop. For each ensemble model $f_i$, a separate perturbation vector $perturb_i$ is optimized. Formally, each PGD step updates the latent perturbation as follows:
\[
\begin{aligned}
\text{perturb}_i^{(t+1)}
&=
\Pi_{\mathcal{P}(d)}\Big(
\text{perturb}_i^{(t)}
+
\alpha
\nabla_{\!\text{perturb}_i}
L_i\big( \\
&\qquad
f_i\big(G(x,\text{perturb}_i^{(t)})\big),
y
\big)
\Big)
\end{aligned}
\]

where $\alpha$ denotes the step size and $\text{Proj}_{\mathcal{P}(d)}$ indicates projection back into the valid perturbation space. This refinement ensures the synthetic adversarial images progressively become more challenging while maintaining identity.

In practice, each recognition model provides an independent
feedback signal during perturbation optimization. The resulting
synthesized images obtained from different models are then
aggregated to produce the final adversarial training sample.

\begin{equation}
x_{\text{ens}} =
\min_{i \in \{1,\dots,N\}}
G(x,\text{perturb}_i)
\end{equation}

This element-wise minimum aggregation selects the most
adversarial response across the ensemble-generated samples.

Our implementation uses a single-step PGD update for perturbation initialization with $\epsilon=\alpha=0.314$. This step provides an initial direction that increases recognition difficulty. Subsequently, the perturbation is refined through the ensemble feedback loop, where the synthesis model and face recognition ensemble jointly adjust the perturbation during training.

Thus, while the initial perturbation follows a single-step PGD formulation (similar to FGSM), the overall optimization remains iterative due to the feedback-driven refinement across training iterations.
This design was chosen for computational efficiency while still enabling difficulty-aware adversarial augmentation.

\noindent\textbf{Generation of Refined Synthetic Samples:} After completing the iterative PGD optimization process, the refined latent vector is used to synthesize updated adversarial face images $x_{perturb}$. These refined synthetic images specifically exploit and reveal vulnerabilities of the ensemble models, enhancing their robustness upon training.

\noindent\textbf{Sample Selection and Balanced Batch Formation:} To prevent redundancy and ensure effective training, synthetic adversarial images are systematically selected based on their collective difficulty across the ensemble. The selection criteria involve maximizing the overall ensemble difficulty, effectively focusing on samples posing diverse challenges to different ensemble components. Selected synthetic adversarial images are then methodically combined with real face images to form balanced mini-batches. This balanced training regime prevents overfitting to synthetic artifacts, ensuring robust generalization to real-world scenarios.

\subsection{Analytical Perspective and Intuition}
The behavior of EFCS can be understood from the perspective of adversarial robustness and distributional generalization.
In standard empirical risk minimization (ERM), face recognition models minimize the expected loss over samples drawn from a data distribution $\mathcal{D}$. 
EFCS introduces controlled adversarial perturbations during synthesis, effectively inducing a perturbed training distribution $\mathcal{D}_{\text{perturb}}$.

This perturbed distribution can be expressed as
\[
\mathcal{D}_{\text{perturb}}(x,y)
= \mathbb{E}_{x' \sim \mathcal{P}(d)}
\left[\mathbf{1}_{\{x' = G(x_{\text{perturb}}^{d})\}} \cdot \mathcal{D}(x,y)\right].
\]

The corresponding training objective under EFCS becomes
\[
\min_{f} \;
\mathbb{E}_{(x,y)\sim \mathcal{D}_{\text{perturb}}}
\left[ L_{\text{ens}}(f(x),y) \right],
\]
which encourages the learned representation to remain stable under structured identity-preserving perturbations.

Under mild assumptions such as continuity of the loss function, smoothness of the data manifold, and bounded perturbations, this objective can be interpreted as a robust optimization problem of the form
\[
\min_{f} \;
\max_{d \sim \mathcal{P}(d)}
\mathbb{E}_{(x,y)\sim \mathcal{D}}
\left[ L_{\text{ens}}\big(f(G(x_{\text{perturb}}^{d})),y\big) \right].
\]

Because the latent perturbation space is constrained to generate realistic variations, the resulting adversarial samples remain semantically meaningful and avoid trivial solutions. 
Overall, EFCS improves robustness by explicitly optimizing for challenging yet identity-consistent synthetic examples, thereby enhancing generalization to real-world conditions.

\noindent\textbf{Ensemble feedback justification.}
Aggregating gradients across multiple recognition models reduces bias toward model-specific failure modes and provides a smoother estimate of the difficulty landscape, consistent with prior findings on ensemble robustness~\cite{ensemble}. 
This promotes perturbations that generalize across architectures and loss formulations, preventing over-specialization to single-model adversarial artifacts and improving the training utility of synthesized samples~\cite{demodalize_synthetic}.

\section{Experimental Details}

\subsection{Evaluation Metrics}

We evaluate the quality, realism, and utility of generated face images using several metrics directly connected to our EFCS objectives. Specifically, we report Fréchet Inception Distance (FID), Kernel Inception Distance (KID), precision, recall, Structural Similarity Index Measure (SSIM), Learned Perceptual Image Patch Similarity (LPIPS), and Deep Image Structure and Texture Similarity (DISTS).

\begin{table}[t]
\centering
\caption{Dataset statistics used for EFCS training. Synthetic images are generated per identity using the EFCS synthesis model.}
\setlength{\tabcolsep}{4pt}
\small
\begin{tabular}{lccc}
\toprule
Dataset & Identities & Images/ID & Total Images \\
\midrule
CASIA-WebFace (original) & 10,575 & $\sim$47 avg & 494,414 \\
Real-only subset & 10,575 & 20 & 211,500 \\
EFCS Synthetic & 10,575 & 20 & 211,500 \\
Mixed (Real + EFCS) & 10,575 & 40 & 423,000 \\
\bottomrule
\end{tabular}

\label{tab:dataset_stats}
\end{table}
\subsection{Implementation Details}

\noindent\textbf{Datasets and preprocessing.}
Experiments are conducted on CASIA-WebFace and WiderFace-12K. All face images are aligned using five landmarks and resized to $112\times112$. 
From the original CASIA-WebFace dataset (494,414 images, 10,575 identities), we construct balanced subsets by selecting 20 real images per identity and generating 20 EFCS samples. 
This yields real-only and EFCS datasets of 211,500 images each, and a mixed dataset of 423,000 images.

\noindent\textbf{Recognition models.}
We evaluate robustness using three representative FR models: ArcFace~\cite{ArcFace}, AdaFace~\cite{adaface}, and ElasticFace~\cite{ElasticFace}, each employing distinct loss formulations.

\noindent\textbf{Controllable synthesis.}
The synthesis network $G$, adapted from~\cite{CFSM}, produces identity-preserving adversarial variations conditioned on a latent perturbation vector and perturbation depth parameter.

\noindent\textbf{Perturbation generation.}
Adversarial samples are generated using single-step PGD with perturbation bound $\epsilon=0.314$ and step size $\alpha=0.314$.

\noindent\textbf{Training setup.}
Models are trained with an ArcFace-ResNet50 backbone using SGD (learning rate $0.1$, momentum $0.9$, weight decay $5\times10^{-4}$) for 20 epochs with polynomial decay and a 2-epoch warmup.

\noindent\textbf{Hardware.}
All experiments are performed on a single NVIDIA A40 GPU (48\,GB).

\section{Results and Analysis}
To validate the proposed EFCS, we performed extensive quantitative evaluations on synthesized face images using a diverse array of metrics, chosen specifically to align with our analytical motivation and optimization goals. 
\begin{table}[t]
\centering
\caption{Average SER-FIQ comparison across synthesis strategies (higher is better).}
\setlength{\tabcolsep}{4pt}
\small
\begin{tabular}{lc}
\toprule
Sample Type & Avg.\ SER-FIQ \\
\midrule
Original & 0.7776 \\
Synthesized & 0.7338 \\
Single Feedback~\cite{CFSM} & 0.7264 \\
EFCS (ours) & 0.7277 \\
\bottomrule
\end{tabular}
\label{tab:serfiq_comparison}
\end{table}
Table~\ref{tab:serfiq_comparison} reports the average SER-FIQ scores for different synthesis strategies. 
SER-FIQ reflects face image quality through embedding robustness and provides an indication of visual consistency in synthesized samples. 
Original images obtain the highest score (0.7776), while random-noise synthesis shows a modest reduction (0.7338). 
EFCS samples (0.7277) yield scores close to those of single-feedback synthesis (0.7264), suggesting broadly comparable perceptual quality. 
These results indicate that ensemble-guided perturbations introduce additional recognition difficulty while largely preserving identity-related visual cues. 

\noindent\textbf{Interpreting Metrics under Hard-Sample Synthesis.}
EFCS is designed to generate identity-preserving hard samples rather than visually identical replicas of real faces. 
Accordingly, metric shifts such as lower SSIM or higher FID/KID should be viewed as indicators of controlled distributional expansion and increased recognition difficulty, rather than reduced synthesis quality.

Table~\ref{tab:metrics_results} supports this interpretation. 
EFCS samples show lower SSIM, reflecting reduced trivial structural similarity, while higher LPIPS (0.5922) and DISTS (0.3653) indicate perceptually meaningful variations introduced through ensemble-guided perturbations. 
These trends are consistent with the objective of producing challenging yet coherent training examples.

Distributional statistics in Table~\ref{tab:metrics_values} further highlight this behavior. 
Compared with single-feedback and random synthesis, EFCS yields higher FID (73.947) and KID (0.0152), reflecting intentional expansion toward harder and under-represented regions of the face manifold. 
Despite this shift, precision remains relatively high (0.7566), suggesting that generated samples retain substantial realism and identity consistency. 
Lower recall (0.3152) indicates that EFCS does not aim for uniform coverage of the real distribution, but instead concentrates on difficult regions that expose model vulnerabilities. 
This targeted diversity aligns with the goal of adversarial augmentation, where training utility is prioritized over distribution completeness.

Finally, Table~\ref{tab:custom_eval} evaluates models trained on real, mixed, and EFCS-only datasets. 
Models trained solely on EFCS achieve strong performance on the EFCS test set (89.6\% R1) and transfer competitively to real data (84.3\% R1). 
Combining EFCS with real data yields the most robust models overall (e.g., 99.4\% R1 on Mix and 99.2\% on Real for Mixed-50\%), whereas real-only training generalizes poorly to synthetic domains (73.1\% R1 on EFCS). Importantly, unlike conventional generative modeling where lower FID/KID universally indicates better realism, EFCS intentionally expands the synthetic distribution toward harder and under-represented regions of the face manifold.
An interesting finding is the performance gap between sequential finetuning, and mixed training. Sequential finetuning exposes the model to one domain at a time, causing strong representation drift and partial forgetting of previously learned features. When training first on EFCS (hard, perturbed samples) and then switching to real data or vice versa, the optimizer adapts to the most recent distribution, often overwriting features that were useful for the earlier domain. In contrast, mixed training provides simultaneous exposure to both real and hard synthetic variations, encouraging the model to learn domain-invariant and more stable identity cues. This joint optimization prevents collapse toward a single distribution and leads to better robustness and generalization. These results demonstrate that EFCS-generated data provides effective standalone supervision and further enhances generalization when used alongside real data.


\begin{table}[t]
\centering
\caption{Perceptual similarity comparison. EFCS produces harder samples, hence lower SSIM and higher LPIPS/DISTS.}
\small
\setlength{\tabcolsep}{4pt}
\begin{tabular}{lrrr}
\toprule
Method & SSIM & LPIPS & DISTS \\
\midrule
Baseline (no feedback) & 0.4016 & 0.4859 & 0.3241 \\
Single feedback~\cite{CFSM} & 0.3859 & 0.5020 & 0.3383 \\
EFCS (ours) & 0.2376 & 0.5922 & 0.3653 \\
\bottomrule
\end{tabular}
\label{tab:metrics_results}
\end{table}

\begin{table}[t]
\centering
\caption{Distributional evaluation (FID↓, KID↓, Prec↑, Rec↑).}
\setlength{\tabcolsep}{4pt}
\small
\begin{tabular}{lrrrr}
\toprule
Sample & FID & KID & Prec. & Rec. \\
\midrule
Synthesized & 57.395 & 0.0123 & 0.7266 & 0.5136 \\
Single fb.~\cite{CFSM} & 63.837 & 0.0132 & 0.8121 & 0.4501 \\
EFCS (ours) & 73.947$\uparrow$ & 0.0152$\uparrow$ & 0.7566 & 0.3152 \\
\bottomrule
\end{tabular}
\label{tab:metrics_values}
\end{table}

\begin{table}[t]
\centering
\caption{Core evaluation on custom test sets (R1/R5 and TAR@\{1e-5,1e-4,1e-3\}, \%).}
\setlength{\tabcolsep}{4pt}
\small
\begin{tabular}{p{2.2cm} l r r r r r}
\toprule
Model & Data & R1 & R5 & T@1e-5 & T@1e-4 & T@1e-3 \\
\midrule
\multirow{3}{*}{EFCS-only}
 & EFCS & 89.6 & 93.0 & 52.4 & 69.0 & 80.5 \\
 & Mix  & 95.3 & 96.7 & 49.2 & 63.7 & 76.0 \\
 & Real & 84.3 & 89.1 & 41.8 & 58.2 & 71.2 \\
\midrule
\multirow{3}{*}{Mixed-50\%}
 & EFCS & 98.9 & 99.3 & 94.9 & 97.0 & 98.2 \\
 & Mix  & 99.4 & 99.6 & 95.9 & 97.4 & 98.5 \\
 & Real & 99.2 & 99.4 & 96.3 & 97.8 & 98.7 \\
\midrule
\multirow{3}{*}{Mixed-33\%}
 & EFCS & 89.0 & 92.8 & 52.5 & 69.0 & 80.3 \\
 & Mix  & 96.1 & 97.2 & 69.8 & 78.8 & 86.1 \\
 & Real & 94.9 & 96.2 & 80.2 & 86.4 & 90.6 \\
\midrule
\multirow{3}{*}{Mixed-25\%}
 & EFCS & 84.4 & 89.9 & 40.9 & 59.1 & 84.4 \\
 & Mix  & 95.4 & 96.7 & 61.3 & 72.8 & 95.4 \\
 & Real & 94.0 & 95.6 & 76.2 & 83.8 & 94.0 \\
\midrule
\multirow{3}{*}{Real-only}
 & EFCS & 73.1 & 82.3 & 25.2 & 41.5 & 57.9 \\
 & Mix  & 94.5 & 96.2 & 50.0 & 63.1 & 74.8 \\
 & Real & 94.5 & 95.8 & 74.5 & 84.1 & 89.5 \\
\midrule
\multirow{3}{*}{Pretr.+EFCS}
 & EFCS & 96.4 & 97.2 & 88.4 & 91.7 & 94.5 \\
 & Mix  & 97.9 & 98.4 & 82.3 & 87.4 & 97.9 \\
 & Real & 92.8 & 94.2 & 73.2 & 82.3 & 87.5 \\
\midrule
\multirow{3}{*}{EFCS→Real ft.}
 & EFCS & 54.1 & 65.0 & 12.3 & 22.4 & 34.9 \\
 & Mix  & 88.5 & 91.4 & 21.7 & 32.3 & 45.3 \\
 & Real & 74.2 & 81.4 & 27.2 & 40.8 & 54.8 \\
\midrule
\multirow{3}{*}{Real→EFCS ft.}
 & EFCS & 65.1 & 74.7 & 19.0 & 31.4 & 45.7 \\
 & Mix  & 86.7 & 90.0 & 16.9 & 27.7 & 41.3 \\
 & Real & 59.2 & 68.1 & 10.6 & 22.6 & 36.1 \\
\bottomrule
\end{tabular}
\label{tab:custom_eval}
\end{table}

\subsection{Graphical Analysis and Detailed Observations}

\begin{figure*}[t]
\centering

\begin{subfigure}[t]{0.32\textwidth}
\centering
\includegraphics[width=\linewidth]{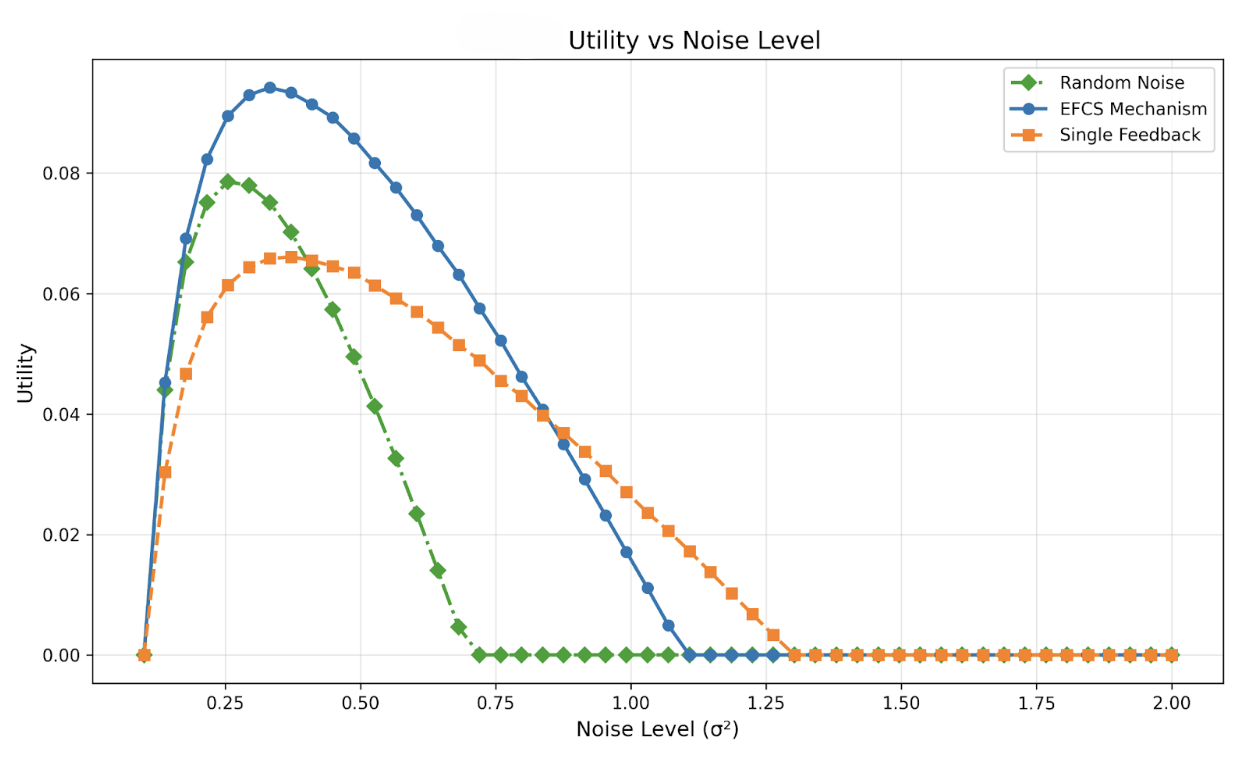}
\caption{Utility vs noise variance $\sigma^2$.}
\end{subfigure}
\hfill
\begin{subfigure}[t]{0.32\textwidth}
\centering
\includegraphics[width=\linewidth]{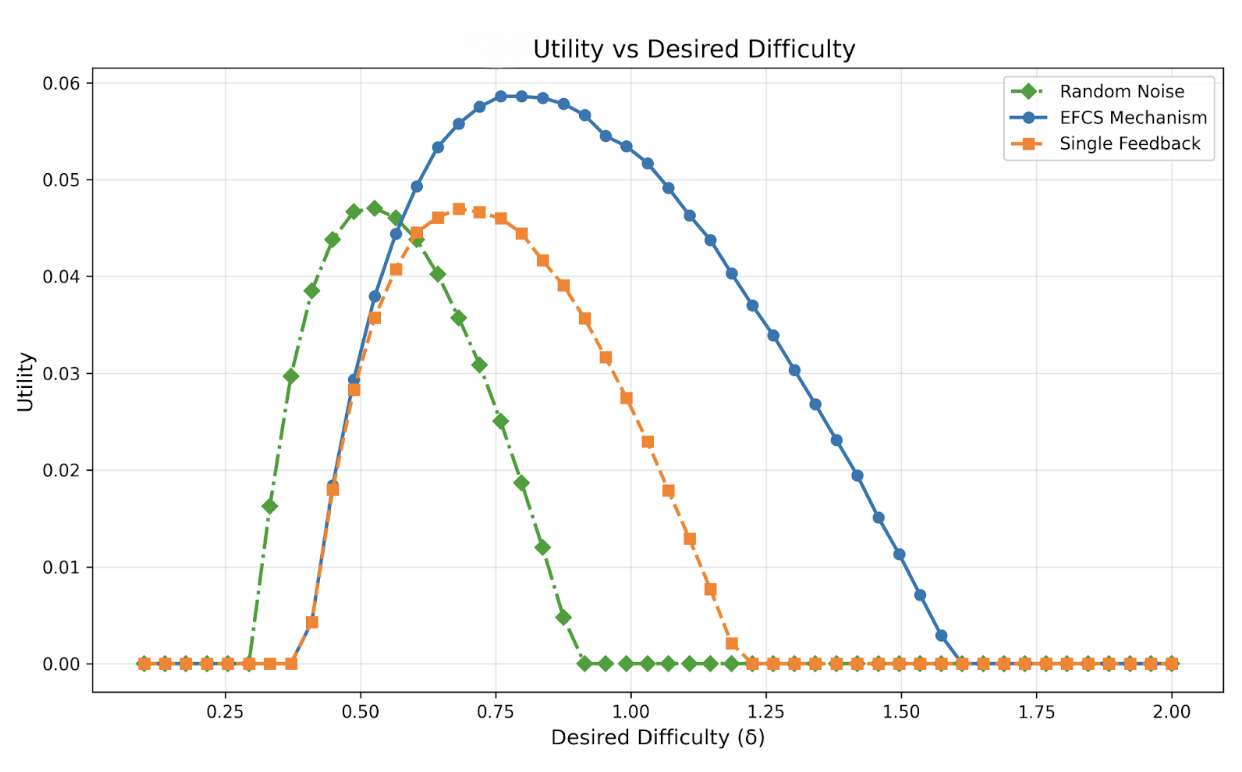}
\caption{Utility vs difficulty $\delta$.}
\end{subfigure}
\hfill
\begin{subfigure}[t]{0.32\textwidth}
\centering
\includegraphics[width=\linewidth]{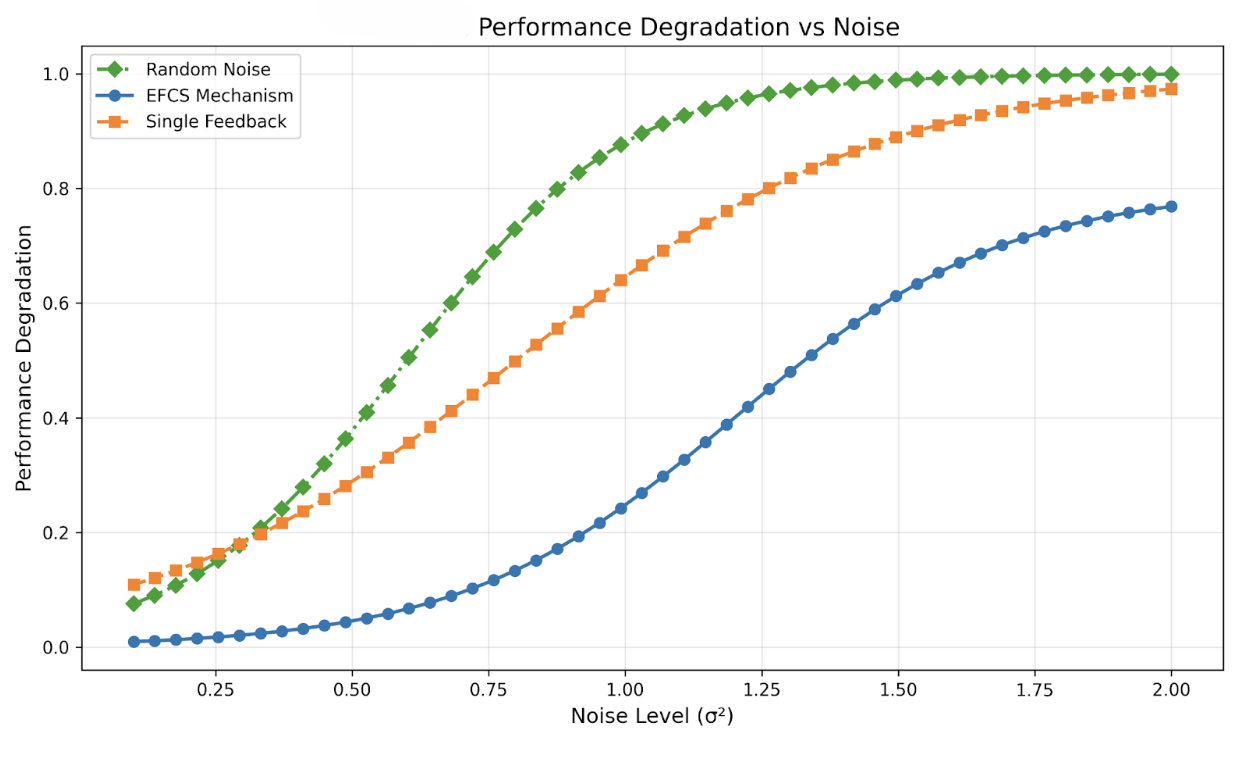}
\caption{Performance degradation vs $\sigma^2$.}
\end{subfigure}

\caption{Effect of synthesis noise and difficulty on training utility and recognition performance. Results averaged across ArcFace, AdaFace, and ElasticFace on CASIA-WebFace. Moderate perturbation improves utility, while excessive noise causes degradation.}
\label{fig:utility_analysis}
\end{figure*}
Figure \ref{fig:utility_analysis} analyzes the effect of synthesis noise and target difficulty on training utility and recognition robustness. Utility exhibits a clear unimodal trend with respect to perturbation strength: very small noise provides limited benefit, while moderate perturbations generate informative hard samples that maximize utility. Beyond a critical noise level, excessive distortion reduces both utility and recognition performance. A similar bell-shaped relationship is observed for the desired difficulty parameter, confirming that intermediate difficulty yields the most effective supervision. The degradation curve further shows that recognition performance remains stable under mild perturbations but drops sharply after a threshold, indicating the transition from beneficial hard samples to unrealistic samples. Overall, these results validate the EFCS design objective of controllably balancing sample difficulty and identity preservation.

\subsection{Human Evaluation Study}
To complement the quantitative analysis, we conducted an ethics-approved human study with 63 participants of varying expertise in AI and computer vision (12.7\% advanced, 38.1\% intermediate, and 49.2\% beginners). The study aimed to assess perceptual similarity and identity recognizability of synthesized faces generated using different feedback mechanisms.

Participants were shown reference images of a real identity alongside a generated sample under three synthesis settings: (i) random perturbation
without feedback, (ii) single-model feedback (1FR), and (iii) the proposed ensemble feedback mechanism (3FR). They rated identity similarity on a 5-point Likert scale (1: very dissimilar, 5: highly similar) and additionally
indicated whether the generated image corresponded to the same identity (\{Yes, No, Not sure\}). To examine perceived difficulty, participants also selected the most challenging yet recognizable sample from sets with progressively increasing noise. To minimize bias, image order and synthesis conditions were randomized, and participants were not informed of the underlying generation method. Responses were aggregated to compute average similarity scores and recognizability statistics.

Across expertise levels, EFCS samples were consistently perceived as more challenging than those produced by random perturbations or single-feedback synthesis while remaining identifiable. Advanced users typically assigned lower similarity scores (2.0–3.0), reflecting increased difficulty with preserved identity cues, whereas intermediate and beginner users reported slightly higher scores. Notably, about 65\% of participants selected samples with intermediate noise levels as the hardest recognizable cases, suggesting that EFCS effectively controls perturbation strength to increase difficulty without compromising identity information. These findings support the intended objective of EFCS: generating identity-preserving yet challenging variations that can enhance the robustness of face recognition models.
\begin{table}[]
\centering
\caption{Training settings of evaluated models.}
\setlength{\tabcolsep}{4pt}
\small
\begin{tabular}{p{2.6cm} p{4.6cm}}
\toprule
Model & Training strategy \\
\midrule
M1 (EFCS) & Scratch on EFCS \\
M2 (Mix-50) & Scratch on EFCS+Real (50\%) \\
M3 (Mix-33) & Scratch on EFCS+Real (33\%) \\
M4 (Mix-25) & Scratch on EFCS+Real (25\%) \\
M5 (Real) & Scratch on Real \\
M6 (Pretr.+EFCS) & Pretrained → finetune EFCS \\
M7 (EFCS→Real) & EFCS model → finetune Real \\
M8 (Real→EFCS) & Real model → finetune EFCS \\
\bottomrule
\end{tabular}
\label{tab:model_specs}
\end{table}
\begin{table*}[]
\centering
\caption{Verification performance (\%) on six public benchmarks. Metrics: accuracy (Acc) and TAR@FAR.}
\setlength{\tabcolsep}{3.2pt}
\small
\begin{tabular}{l ccc ccc ccc ccc ccc ccc}
\toprule
& \multicolumn{3}{c}{LFW}
& \multicolumn{3}{c}{CFP-FP}
& \multicolumn{3}{c}{AgeDB}
& \multicolumn{3}{c}{CALFW}
& \multicolumn{3}{c}{CPLFW}
& \multicolumn{3}{c}{CFP-FF} \\
\cmidrule(lr){2-4}\cmidrule(lr){5-7}\cmidrule(lr){8-10}
\cmidrule(lr){11-13}\cmidrule(lr){14-16}\cmidrule(lr){17-19}
Model
& Acc & 1e-4 & 1e-3
& Acc & 1e-4 & 1e-3
& Acc & 1e-4 & 1e-3
& Acc & 1e-4 & 1e-3
& Acc & 1e-4 & 1e-3
& Acc & 1e-4 & 1e-3 \\
\midrule

M1 & 98.03 & 29.81 & 89.52 & 87.10 & 0.07 & 0.73 & 87.95 & 5.76 & 27.22 & 89.90 & 14.90 & 50.80 & 81.97 & 22.50 & 26.20 & 97.49 & 1.90 & 19.02 \\

M2 & 98.68 & 29.15 & 92.33 & 90.57 & 0.18 & 1.80 & 90.70 & 9.56 & 33.13 & 91.28 & 6.02 & 51.70 & 84.65 & 5.65 & 37.70 & 98.51 & 3.25 & 32.53 \\

M3 & 98.65 & 27.98 & 90.78 & 90.27 & 0.23 & 2.30 & 90.73 & 10.22 & 38.37 & 91.40 & 14.34 & 57.87 & 84.55 & 7.64 & 31.30 & 98.41 & 3.30 & 33.28 \\

M4 & 98.83 & 30.39 & 94.97 & 90.21 & 0.52 & 5.20 & 91.00 & 6.04 & 43.03 & 92.03 & 15.51 & 54.53 & 84.20 & 6.32 & 29.18 & 98.29 & 3.40 & 33.95 \\

M5 & 98.88 & 30.72 & 94.60 & 91.33 & 0.38 & 3.80 & 90.98 & 13.70 & 48.63 & 92.08 & 15.60 & 62.00 & 85.03 & 6.35 & 39.10 & 98.41 & 3.60 & 36.02 \\

M6 & 99.38 & 36.60 & 98.60 & 95.17 & 0.33 & 3.30 & 93.85 & 13.90 & 52.90 & 93.38 & 20.60 & 68.10 & 88.33 & 11.80 & 46.20 & 99.14 & 4.47 & 44.70 \\

M7 & 95.58 & 19.70 & 74.90 & 76.69 & 0.01 & 0.12 & 80.83 & 2.48 & 12.60 & 84.62 & 4.67 & 26.30 & 75.00 & 4.30 & 15.10 & 94.39 & 0.97 & 9.67 \\

M8 & 93.63 & 16.90 & 61.90 & 72.23 & 0.00 & 0.00 & 69.43 & 0.29 & 1.90 & 81.12 & 2.81 & 16.70 & 71.02 & 1.13 & 8.20 & 91.17 & 0.55 & 5.53 \\

\bottomrule
\end{tabular}
\label{tab:benchmarks}
\end{table*}
\subsection{Model Overview}
To evaluate the impact of synthetic data on facial recognition model performance, we developed eight distinct models, summarized in Table \ref{tab:model_specs}. All models share a common architectural foundation, utilizing an iResNet-50 backbone with an ArcFace head to ensure a consistent basis for comparison. Our experiments explore various training methodologies: some models were trained from scratch using different datasets, while others were fine-tuned from pre-existing weights.

The core of our investigation lies in the 'Mixed eval' models (M2, M3, M4), which were trained on hybrid datasets containing both real and synthetic images. The percentage specified in their names: 50\%, 33\%, and 25\% denotes the proportion of synthetic data included in their respective training sets. This allows for a direct analysis of how the quantity of synthetic data influences model accuracy and robustness. The remaining models serve as crucial baselines, including models trained exclusively on synthetic data (M1), exclusively on real data (M5), and fine-tuned variants (M6, M7, M8) to assess the efficacy of transfer learning.

\subsection{Analysis of Public Benchmark Performance}
The performance of our models on six standard public benchmarks is detailed in Table \ref{tab:benchmarks}. A key finding from these results is the competitive performance of the models trained with a combination of real and synthetic data (M2, M3, and M4) when compared to the model trained exclusively on real-world data (M5). Across all benchmarks, including LFW, CFP-FP, and AgeDB-30, the 'Mixed eval' models consistently achieve metrics that are only marginally lower than the 'Real eval' model. For instance, on the LFW benchmark, the accuracy of M2 (98.68\%) and M4 (98.83\%) is nearly identical to that of M5 (98.88\%).

While the model trained on pure real data (M5) sets a high-performance baseline, the minimal difference in performance demonstrates the viability of synthetic data as a powerful supplement or even an alternative to real data. This slight trade-off in accuracy is highly acceptable when considering the immense benefits of synthetic data generation: the ability to create unlimited, diverse, and unbiased datasets on demand. This approach not only mitigates critical data privacy and security concerns associated with collecting real images but also offers a scalable solution to overcome data scarcity and systematically reduce the demographic biases often present in real-world datasets.

\section{Conclusion}
The proposed Ensemble Feedback Controllable Synthesis method advances synthetic data generation for face recognition by effectively balancing complexity with recognizability. EFCS surpasses traditional methods by creating synthetic samples specifically designed to enhance model robustness and generalization. Quantitative evaluations using metrics such as SSIM, LPIPS, and DISTS confirm EFCS's capability to generate challenging yet authentic samples. The method addresses limitations in existing synthetic datasets and provides an analytical motivation for optimized noise injection. By producing controlled synthetic data that closely emulates real-world conditions, EFCS offers a robust and privacy-conscious alternative for training FR models. 


\printbibliography
\end{document}